\documentclass[preprint]{article}

\usepackage[nonatbib]{neurips_2023}

\usepackage{array}
\usepackage{longtable}
\usepackage{caption}
\usepackage[table]{xcolor}
\usepackage[utf8]{inputenc} 
\usepackage[T1]{fontenc}    
\usepackage{hyperref}       
\usepackage{url}            
\usepackage{booktabs}       
\usepackage{amsfonts}       
\usepackage{amsmath}
\usepackage{nicefrac}       
\usepackage{microtype}      
\usepackage{xcolor}         
\usepackage{graphicx}
\usepackage{tikz}
\usepackage{multicol}
\usepackage{multirow}

\usepackage{algorithm}
\usepackage{algpseudocode}
\usepackage{tabularx} 
\title{SpikeMba: Multi-Modal Spiking Saliency Mamba for Temporal Video Grounding}
\usepackage{subcaption}

\author{%
  Wenrui Li$^{1}$ \qquad Xiaopeng Hong$^{1}$ \qquad Ruiqin Xiong$^{2}$ \qquad Xiaopeng Fan$^{1}$\\
  $^{1}$ Harbin Institute of Technology \\
  $^{2}$ Peking University\\
  \{liwr618@163.com; hongxiaopeng@ieee.org; rqxiong@pku.edu.cn; fxp@hit.edu.cn\} \\
}

\begin{document}

\maketitle

\begin{abstract}
Temporal video grounding (TVG) is a critical task in video content understanding, requiring precise alignment between video content and natural language instructions. Despite significant advancements, existing methods face challenges in managing confidence bias towards salient objects and capturing long-term dependencies in video sequences. To address these issues, we introduce SpikeMba: a multi-modal spiking saliency mamba for temporal video grounding. Our approach integrates Spiking Neural Networks (SNNs) with state space models (SSMs) to leverage their unique advantages in handling different aspects of the task. Specifically, we use SNNs to develop a spiking saliency detector that generates the proposal set. The detector emits spike signals when the input signal exceeds a predefined threshold, resulting in a dynamic and binary saliency proposal set. To enhance the model's capability to retain and infer contextual information, we introduce relevant slots—learnable tensors that encode prior knowledge. These slots work with the contextual moment reasoner to maintain a balance between preserving contextual information and exploring semantic relevance dynamically. The SSMs facilitate selective information propagation, addressing the challenge of long-term dependency in video content. By combining SNNs for proposal generation and SSMs for effective contextual reasoning, SpikeMba addresses confidence bias and long-term dependencies, thereby significantly enhancing fine-grained multimodal relationship capture. Our experiments demonstrate the effectiveness of SpikeMba, which consistently outperforms state-of-the-art methods across mainstream benchmarks.

\end{abstract}

\section{Introduction}
The widespread adoption of the internet and smart devices has dramatically increased the production and consumption of video data. Concurrently, demand for retrieving and understanding video content has grown. Within this context, video temporal localization has emerged as a key research area. The main challenge in video temporal localization involves understanding the complex relationships between video content and natural language instructions, and pinpointing the exact segments that match a given description. Addressing this challenge depends on efficiently aligning and understanding multimodal spatio-temporal features. Lin et al. \cite{UniVTG} develop a unified framework with scalable pseudo supervision, establishing an effective grounding model to benefit temporal grounding pretraining with various labels. Liu et al. \cite{liucvpr15} introduce the unified multimodal transformers, marking the first framework that allows for the joint optimization of moment retrieval and highlight detection in videos through multimodal learning. Additionally, several studies \cite{QD-DETR,cg-detr} have focused on the semantic relevance between video clips and textual queries.

Although current methods in video content localization show significant performance, they still face challenges with several key issues: \textbf{1). Confidence Bias towards Salient Objects:} In complex video environments, models tend to identify multiple potential proposal sets because of drastic changes in salient objects. This results in an excessive focus on these objects, often neglecting the video's overall content. For instance in Fig. 1 (left), the highest confidence proposal (prediction \#1) might focus solely on a door opening, ignoring the subsequent action of a dog entering. This suggests that a secondary proposal (prediction \#2), which might better reflect the video content, often receives a lower prediction score despite being closer to the correct interpretation. \textbf{2). Long-Term Dependency of Relevant Segments:} In video temporal localization, especially with long sequences, it is crucial to address complex long-term spatial-temporal relationships. Although Transformer models demonstrate advantages in capturing long-term dependencies, they tend to engage in uniform cross-modal interactions between every word and each video clip for each query. This would overlooks the selective focus on relevant video segments, potentially ignoring attention towards the segments critical for understanding the video's overall action.

To tackle the aforementioned challenges, we propose a novel network architecture, as illustrated in Fig. 1 (right). Our model concentrates on two main aspects: \textbf{1). Saliency Proposal Set:} We utilize Spiking Neural Networks (SNNs) to build an advanced saliency detector. This detector aims to explore the deep connections between features of different modalities and capture key proposal clues. SNNs use a unique threshold mechanism to generate a saliency proposal set. They emit a spike signal when the input signal strength exceeds a predefined threshold, producing a binary sequence that forms the saliency proposal set. Importantly, the saliency proposals generated by the SNN vary across different time steps, with each time step representing a different confidence level of potential proposals. By increasing the number of time steps, we can enhance the proposal suggestions, thereby improving the model's comprehensive understanding of video content and overcoming confidence bias towards salient objects. \textbf{2). Relevant Prior Knowledge:} Due to the limitations of memory capacity, long videos typically need to be sampled at a fixed rate before processing by the model. To optimize this process, we introduce learnable tensors, known as Relevant Slots, to encode learned prior knowledge. These slots enhances the model's deep understanding of video content by maintaining contextual information over long sequences.
\begin{figure}
    \centering
	\includegraphics[scale=0.5]{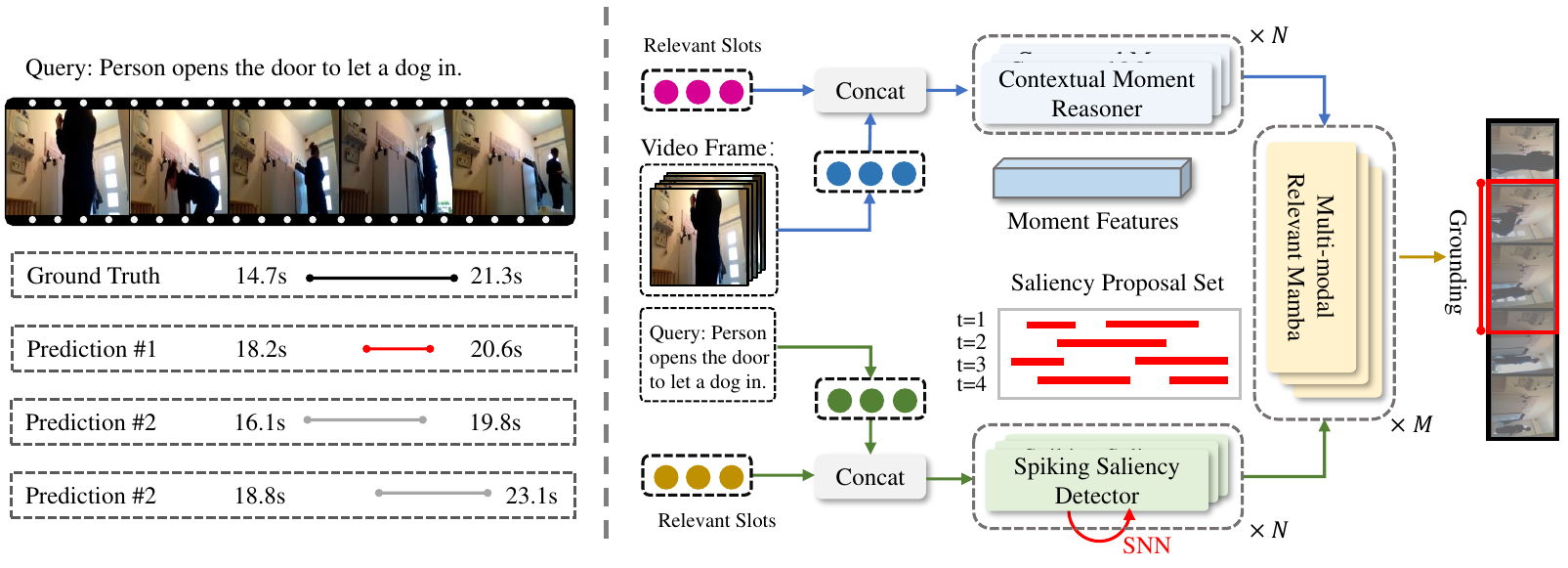}
	\caption{An illustration of the challenges in video grounding task (left) and the overall architecture of our proposed model (right). On the left, a scenario underscores the problem of confidence bias towards salient objects, with models overemphasizing dramatic changes. On the right, we employ the SNN and Mamba for dynamic proposal generation, encoding prior knowledge, and improving understanding of long-sequence videos.}
	\label{fig:5}
\end{figure}

In this paper, we present a multi-modal spiking saliency mamba for temporal video grounding. Specifically, the set of input visual and textual features is extended with additional "relevants slots" to encode the priori knowledge for contextual information. The contextual moment reasoner leverages these learnable slots to balance preserving contextual information and exploring semantic relevance. To capture saliency proposal clues more effectively, we introduce a spiking saliency detector. This detector uses the threshold mechanism of SNN and generated binary sequences to explore potential saliency proposals. The SNN's dynamic adjustment of time steps helps in focusing on various saliency proposals. Finally, we introduce the multi-modal relevant mamba block based on state space models (SSMs). The dynamic parameterization of SSMs based on input enables the model to selectively propagate or forget information, benefiting our proposed relevant slots. By efficiently coupling these modules, our model demonstrates its superiority in handling continuity and context dependency in video content, significantly improving the accuracy and efficiency of video content localization.

To sum up, the main contributions of this paper are as follows:

1) We present a novel SpikeMba: multi-modal spiking saliency mamba for temporal video grounding. We propose a spiking saliency detector, which utilize the threshold mechanism of SNN and generated binary sequences to effectively explore potential saliency proposals.

2) To improve the memory capabilities for contextual information across long video sequences, we introduce the relevant slots to selectively encode prior knowledge. The contextual moment reasoner is proposed to further inference the contextual semantic associations with dynamically relevant slots.

3) We incorporate the SSMs to selective information propagation based on the current input, effectively addressing the challenge of long-term dependency in video content. Our experiments demonstrate the effectiveness of SpikeMba in benchmark datasets.

\section{Related Work}
\textbf{Video Temporal Grounding} involves predicting and accurately locating the segments within a video that match a given description based on input videos and natural language queries. Models based on the Transformer architecture are widely applied in the joint understanding of video and language\cite{UniVTG,QD-DETR,Moment-DETR,EaTR,LLAVILO,LIM,joint-va}. Zhang et al. \cite{tvg1} explore the integration of text and visual prompts with the CLIP model to enhance temporal video grounding using 2D visual features, addressing the gap in multi-modal learning. Pan et al. \cite{tvg2} present approaches for temporal grounding in long videos, tackling the challenges of localizing moments within extensive footage. Soldan et al. \cite{tvg3} introduce a graph-matching network aimed at facilitating mutual information exchange across video and query domains, focusing on temporal language localization in videos. However, the significant computational complexity of Transformer processing long video sequences ($O(n^{2})$) is a key limitation that restricts the exploration of their long-time dependency capabilities.

Recent efforts have increasingly focused on the completeness of proposal selection for temporal moment localization. Rodriguez et al. \cite{tvg4} discuss proposal-free methods which offering a more efficient and flexible approach for practical applications. Hao et al. \cite{tvg5} emphasize the critical role of negative sample selection in temporal grounding and propose a novel framework to tackle temporal bias effectively. Kim et al. \cite{tvg6} delve into zero-shot learning for video temporal grounding, utilizing narration as queries to guide the learning process in the absence of direct language input. However, the sudden changes in scenes can still compromise the robustness and accuracy of proposal extraction.

\textbf{Spiking Neural Networks} have inherent ability to efficiently process and encode temporal information which benefit in handling video temporal data. This spike-based processing enables SNNs to achieve high temporal precision and energy efficiency since computation occurs only in response to incoming spikes, not continuously. Consequently, SNNs have the potential to attain higher accuracy and lower power consumption than traditional neural networks. Some researches focus on the training strategies and inner mechanisms of SNN \cite{s1,s2,s3,s4,s5}. Li et al. \cite{snn1} propose a novel approach that optimizes time step allocation based on input specificity, enhancing efficiency without compromising accuracy. Jiang et al. \cite{snn2} introduce a pioneering SNN conversion pipeline for Transformers, overcoming limitations of existing methods by approximating complex operations like self-attention and test-time normalization in both temporal and spatial dimensions. Moreover, the research in the field of SNNs for video processing is rapidly evolving. For instance, Ren et al. \cite{snn3} present a novel point-based SNN architecture that efficiently processes sparse event data, by avoiding conventional frame conversion and utilizing a singular-stage structure for feature extraction. Li et al. \cite{snn4} propose a novel dual-stream architecture for audio-visual zero-shot learning, demonstrating the potential of SNNs in handling contextual semantic and dynamic motion information. These aforementioned studies highlight the significant potential of SNNs in advancing the field of video analysis and other time-sensitive applications, offering a promising avenue for research and development.

To address the aforementioned challenges, we leverage state space models \cite{mamba,mamba1,mamba2,mamba3,mamba4,mamba5,mamba6,mamba7,mamba8} to enhance sequence processing efficiency and introduce relevant slots to bolster the model's capability in capturing long-term dependencies. Additionally, we incorporate Spiking Neural Networks (SNNs) as saliency proposal extractors. The threshold mechanism and temporal dynamics of SNNs enable effective identification of salient events within variable video environments while minimizing the impact of noise and sudden changes.

\begin{figure*}
    \centering
	\includegraphics[scale=0.53]{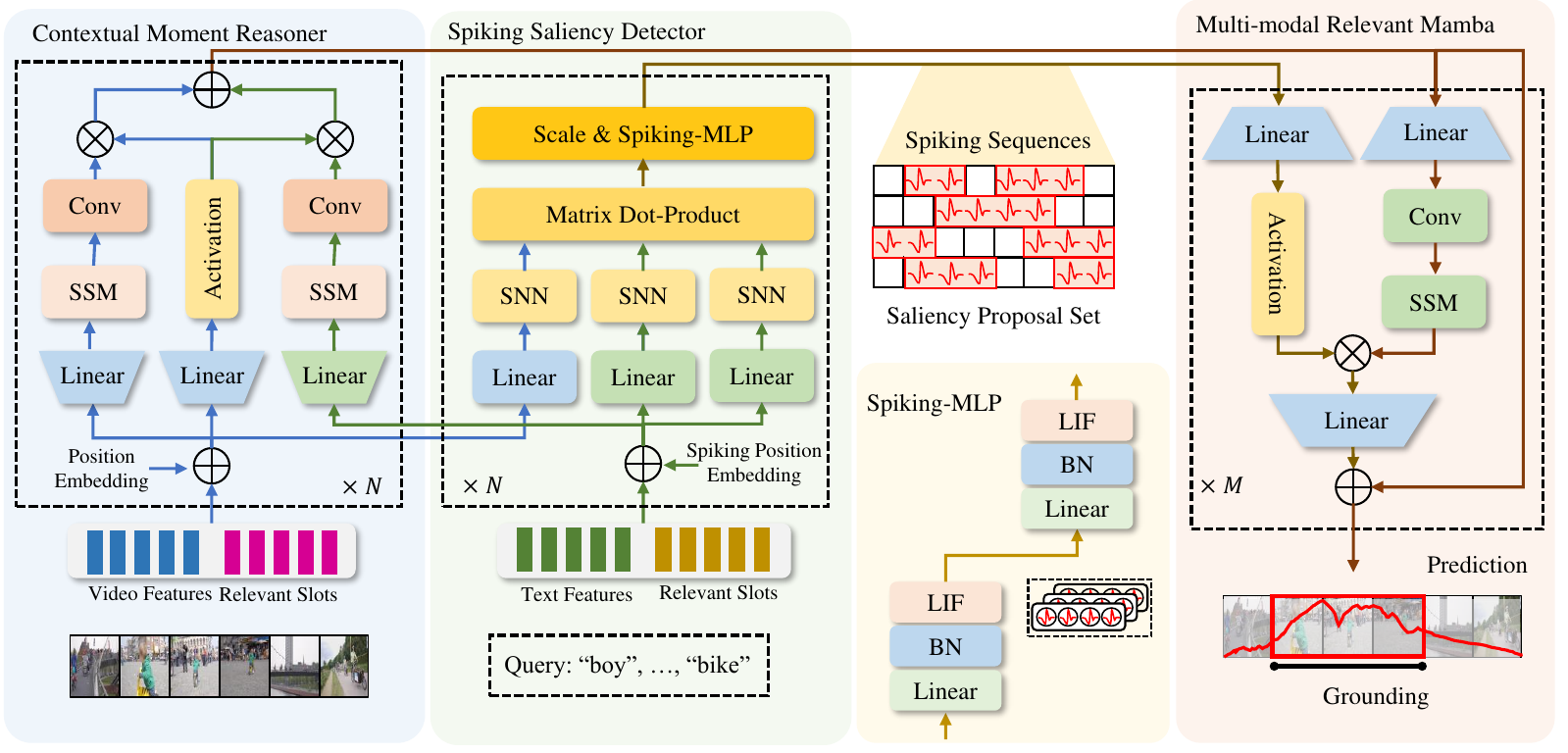}
	\caption{Architectural Overview of the Multi-Modal Spiking Saliency Mamba. The contextual moment reasoner dynamically leverages relevant slots for semantic association and inference. The spiking saliency detector generates a potential proposal set. The multi-modal relevant mamba block enhances long-range dependency modeling while maintaining linear complexity relative to input size. }
	\label{fig:5}
\end{figure*}

\section{Methods}
Given a video $\boldsymbol{V}=[v_{1},\ldots,v_{N_{v}}]$ comprising $N_{v}$ clips, and a textual query $\boldsymbol{Q}=[q_{1},\ldots,q_{N_{q}}]$ containing $N_{q}$ tokens, we design a model to compute the clip-wise saliency scores $[{s_{1},s_{2},\ldots,s_{N_{v}}}]$ and identify target moments as $(\boldsymbol{I}_{c},\boldsymbol{I}_{\omega})$, where $\boldsymbol{I}_{c}$ represents the central temporal coordinate and $\boldsymbol{I}_{\omega}$ represents the duration of the identified moment.
\subsection{State Space Model (SSM)}
State Space Model (SSM) are utilized to describe the evolution of a linear system's state over time, mathematically mapping a function $x(t)\in \mathbb{R} \xrightarrow{} y(t)\in \mathbb{R}$ via a hidden space $h(t)\in \mathbb{R}^{N}$. The SSM in linear system can be mathematically represented as follows:
 \begin{equation}
	\begin{aligned}
        h^{\prime}(t)=\boldsymbol{A}h(t)+\boldsymbol{B}x(t), \quad y(t)=\boldsymbol{C}h(t),
	\end{aligned}
\end{equation}
where $\boldsymbol{A} \in \mathbb{R}^{N\times N}$, $\boldsymbol{B} \in \mathbb{R}^{N\times M}$ and $\boldsymbol{C} \in \mathbb{R}^{M\times N}$ are corresponding to state matrix, input matrix and output matrix. The Mamba model approximate the above continuous system by discretization operation, utilizing a timescale parameter $\delta$ to convert the continuous parameters $\boldsymbol{A}$, $\boldsymbol{B}$ to their discrete counterparts $\boldsymbol{\bar{A}}$, $\boldsymbol{\bar{B}}$. The transformation commonly utilize the zero-order hold (ZOH) method, which can be written as:
 \begin{equation}
	\begin{aligned}
        \boldsymbol{\bar{A}}&=exp(\delta\boldsymbol{A}), \quad
        \boldsymbol{\bar{B}}=(\delta\boldsymbol{A}^{-1}(exp(\delta\boldsymbol{A})-I)\cdot \delta\boldsymbol{B}), \\
        y_{t} &= \boldsymbol{C}h_{t}, \quad \quad \quad h_{t} = \bar{\boldsymbol{A}}h_{t-1}+\bar{\boldsymbol{B}}x_{t}.
	\end{aligned}
\end{equation}

Finally, the output is vectorized into a single convolution:
 \begin{equation}
	\begin{aligned}
        y=x*\boldsymbol{\bar{K}} \quad with \quad \boldsymbol{\bar{K}}=(\boldsymbol{C}\boldsymbol{\bar{B}},\boldsymbol{C}\boldsymbol{\bar{A}}\boldsymbol{\bar{B}}, \ldots,\boldsymbol{C}\boldsymbol{\bar{A}}^{M-1}),   
	\end{aligned}
\end{equation}
where $\boldsymbol{\bar{K}}$ represents the convolution filter and $M$ represents the input sequence length.

\subsection{Contextual Moment Reasoner (CMR)}
We introduce the contextual moment reasoner (CMR) to tackle the challenges of accurately understanding and connecting semantic relationships between video content and textual queries. By dynamically utilizing relevant slots, CMR can infer and balance the context of current moments with their semantic relevance, thereby enhancing the precision and efficiency of video segment localization in diverse and complex environments. Relevant slots are learnable matrices that are concatenated to the original input sequences $\boldsymbol{O}_{l-1}^{vis}$ and $\boldsymbol{O}_{l-1}^{tex}$, enhancing the model's capacity to retain and process long-term dependencies. CMR employs a series of Conv and SSM layers to capture spatial-temporal patterns, followed by linear transformations for dimensionality adjustment. The processed features are then combined with position embeddings to preserve temporal information.

We describe the operations of CMR in Algorithm 1. Initially, CMR requires $\boldsymbol{T}_{l-1}^{vis}$ and $\boldsymbol{T}_{l-1}^{tex}$, corresponding to visual and textual token sequences. The visual and textual relevant slots $\boldsymbol{R}^{vis}$ and $\boldsymbol{R}^{tex}$ are concatenated to these sequences as $\boldsymbol{T}_{l-1}^{vis} = [\boldsymbol{O}_{l-1}^{vis},\boldsymbol{R}^{vis}]$ and $\boldsymbol{T}_{l-1}^{tex} = [\boldsymbol{O}_{l-1}^{tex},\boldsymbol{R}^{tex}]$.

\definecolor{commentcolor}{rgb}{0.0, 0.6, 0.0} 
\definecolor{keywordcolor}{rgb}{0.0, 0.0, 0.8} 
\definecolor{stringcolor}{rgb}{0.6, 0.0, 0.0} 

\renewcommand{\algorithmiccomment}[1]{\hfill\textit{\color{commentcolor}{// #1}}}
\renewcommand{\algorithmicrequire}{\textbf{\color{keywordcolor}{Input:}}}
\renewcommand{\algorithmicensure}{\textbf{\color{keywordcolor}{Output:}}}

\algdef{SE}[FOR]{For}{EndFor}[1]{\algorithmicfor\ #1\ \algorithmicdo}{\algorithmicend\ \algorithmicfor} 

\begin{algorithm}
\caption{CMR Block}
\begin{algorithmic}[1]
\Require token sequence $\boldsymbol{T}_{l-1}^{vis} : (B,M,C)$, $\boldsymbol{T}_{l-1}^{tex} : (B,M,C)$
\Ensure token sequence $T_l^{vis} : (B,M,C)$
\State $Z : (B,M,E) \leftarrow \texttt{Linear}^z(\boldsymbol{T}'_{l-1}{})$
\State \algorithmiccomment{normalize the input sequence $\boldsymbol{T}_{l-1}$}
\For{$o \in {vis, tex}$}
\State $\boldsymbol{T}_{l-1}^{'o} : (B,M,C) \leftarrow \texttt{Norm}(\boldsymbol{T}_{l-1}^o)$
\State $x'_o : (B,M,E) \leftarrow \texttt{Linear}^x_o(\boldsymbol{T}_{l-1}{'o})$
\State $x_o' : (B,M,E) \leftarrow \texttt{SiLU}(\texttt{Conv1d}_o(x_o))$
\State $\boldsymbol{B}_o : (B,M,C) \leftarrow \texttt{Linear}\boldsymbol{B}^o(x_o')$
\State $\boldsymbol{C}_o : (B,M,C) \leftarrow \texttt{Linear}\boldsymbol{C}^o(x_o')$
\State \algorithmiccomment{Parameters function}
\State $\boldsymbol{\Delta}_o : (B,M,E) \leftarrow \log(1 + \exp(\texttt{Linear}\boldsymbol{A}^o(x_o') + \texttt{Parameter}\boldsymbol{A}^o))$
\State $\bar{\boldsymbol{A}_o} : (B,M,E,N) \leftarrow \boldsymbol{\Delta}_o \otimes \texttt{Parameter}\boldsymbol{A}^o$
\State $\bar{\boldsymbol{B}_o} : (B,M,E,N) \leftarrow \boldsymbol{\Delta}_o \otimes \boldsymbol{B}_o$
\State $y_o : (B,M,E) \leftarrow \texttt{SSM}(\boldsymbol{A}_o, \boldsymbol{B}_o, \boldsymbol{C}_o)(x_o')$
\EndFor
\State \algorithmiccomment{get gated $y_{vis}$ and $y_{tex}$}
\State $y_{vis} : (B,M,E) \leftarrow y_{vis} \odot \texttt{SiLU}(z)$
\State $y_{tex} : (B,M,E) \leftarrow y_{tex} \odot \texttt{SiLU}(z)$
\State $\boldsymbol{T}_l^{vis} : (B,M,C) \leftarrow \texttt{Linear}^ T(y_{vis} + y_{tex})$
\State \Return $\boldsymbol{T}_l^{vis}$
\end{algorithmic}
\end{algorithm}

\subsection{Spiking Saliency Detector (SSD)}
We introduce the spiking saliency detector (SSD) to improve the detection of salient moments in videos. The spiking neuron can convert continuous feature sequences into discrete spiking sequences via a threshold mechanism. These spikes effectively represent highly relevant or salient instances within the video, matching the semantic content of the textual query. The dynamics of LIF layer can be written as:
 \begin{equation}
	\begin{aligned}
        \boldsymbol{U}[t] &= \boldsymbol{H}[t - 1] + \boldsymbol{X}[t],\quad \boldsymbol{S}[t] = \text{Hea}(\boldsymbol{U}[t] - U_{\text{th}}),\\
        \boldsymbol{H}[t] &= V_{\text{reset}}\boldsymbol{S}[t] + (\beta \boldsymbol{U}[t])(1 - \boldsymbol{S}[t]), 
	\end{aligned}
\end{equation}
where $t$ is the time step, $\boldsymbol{U}[t]$ is the membrane potential, $\boldsymbol{X}[t]$ denotes the current input, and also the neuron's output at time $t$, while $\text{Hea}(\cdot)$ denotes the Heaviside step function, and $\beta$ scales the contribution of $\boldsymbol{U}[t]$. The spiking sequences generated by the SNN encapsulate the temporal dynamics of salient features in the video. This is achieved through the temporal resolution of spikes, enabling fine-grained analysis of video content over time. The spiking sequences are analyzed to construct the saliency proposal set. Each spike in the sequence is considered a candidate for marking a salient moment, and the collection of spikes forms the basis of the proposal set. This proposal set represents potential moments of interest in the current video which to correspond to the textual query, serving as a crucial input for the grounding process.

The overall architecture can be expressed as follows:
 \begin{equation}
	\begin{aligned}
        \boldsymbol{Q} &= \text{SNN}(\boldsymbol{\mathrm{W}}_{q}\boldsymbol{X}), \boldsymbol{K} = \text{SNN}(\boldsymbol{\mathrm{W}}_{k}\boldsymbol{X}), \boldsymbol{V} = \text{SNN}(\boldsymbol{\mathrm{W}}_{v}\boldsymbol{X}),\\
        \boldsymbol{G} &= \text{Scale}(\boldsymbol{QK}^{T}\boldsymbol{V}), \boldsymbol{S} = \text{SNN}(\text{BN}(\text{Linear}(\boldsymbol{G}))),
	\end{aligned}
\end{equation}
where $\text{SNN}(\cdot)$ denotes the LIF layer, and $\boldsymbol{\mathrm{W}}_{q}$, $\boldsymbol{\mathrm{W}}_{k}$, and $\boldsymbol{\mathrm{W}}_{v}$ are learnable linear matrices, with $\text{Scale}$ being the scaling factor. Leveraging the unique properties of SNNs for saliency detection, the proposed SSD offers a novel and efficient method for identifying key moments in video content. Its ability to generate high-temporal-resolution spiking sequences, along with integration with contextual reasoning mechanisms, ensures accurate localization of moments in video sequences.
\subsection{Multi-modal Relevant Mamba (MRM)}
The Multi-modal Relevant Mamba (MRM) integrates processed video and text features using a combination of linear transformations and convolutional layers. The MRM can predict the temporal locations of moments in the video corresponding to the textual query by analyzing spiking sequences and contextually enriched multimodal features.

The "MRM Block" algorithm processes input token sequences $\boldsymbol{T}_{l-1}$ and $\boldsymbol{S}_{l-1}$, each with dimensions $(B,M,C)$. It employs a series of transformations to enhance their representation for machine learning applications, especially in natural language processing. Initially, both sequences are subjected to layer normalization to stabilize learning. This is followed by linear transformations that modify their feature space dimensions to $(B,M,P)$. The transformed $x$ sequence undergoes further processing through a 1-dimensional convolution and SiLU activation. The UpdateParameters$(\cdot)$ function, similar to that in the CMR block, utilizes the SSM to produce a gated output $y'$ via element-wise multiplication with a gating factor derived from $z$. Incorporating a residual connection, the final output $\boldsymbol{T}_l$ is obtained by adding the transformed $y'$ to the original input $\boldsymbol{T}_{l-1}$. This ensures the algorithm captures and enhances the features of the input sequences while maintaining a connection to the original data, indicative of advanced deep learning techniques designed to capture complex data patterns.
\begin{algorithm}
\caption{MRM Block}
\begin{algorithmic}[1]
\Require token sequence $\boldsymbol{T}_{l-1}: (B,M,C)$, $\boldsymbol{S}_{l-1}: (B,M,C)$
\Ensure token sequence $\boldsymbol{O}_l: (B,M,C)$
\State $\boldsymbol{T}'_{l-1}: (B,M,C) \gets \text{Norm}(\boldsymbol{T}_{l-1})$
\State $\boldsymbol{S}'_{l-1}: (B,M,C) \gets \text{Norm}(\boldsymbol{S}_{l-1})$
\State $x: (B,M,E) \gets \text{Linear}^T(\boldsymbol{T}'_{l-1})$
\State $z: (B,M,E) \gets \text{Linear}^T(\boldsymbol{S}'_{l-1})$
\State $x': (B,M,E) \gets \text{SiLU (Conv1d}(x))$
\State $\bar{\boldsymbol{A}}:(B,M,E,N), \bar{\boldsymbol{B}}:(B,M,E,N), \boldsymbol{C}:(B,M,C) \gets \text{UpdateParameters}(x')$
\State $y: (B,M,E) \gets \text{SSM}(\bar{\boldsymbol{A}},\bar{\boldsymbol{B}},\boldsymbol{C})(x')$
\State $y': (B,M,E) \gets y \circ \text{SiLU}(z)$
\State $\boldsymbol{O}_l: (B,M,C) \gets \text{Linear}^T(y') + \boldsymbol{T}_{l-1}$
\State \textbf{return} $\boldsymbol{O}_l$
\end{algorithmic}
\end{algorithm}
\subsection{Training Strategy}
We trained the SpikeMba model on six Nvidia V100S GPUs. Specifically, we set batch size to 64/64/8/8, learning rate to 2e-4/4e-4/2e-3/4e-4, spiking time step to 8/8/10/10, number of mamba block 6/6/6/6 for QVHighlights, Charades-STA, TVSum and Youtube-HL, respectively. The training is optimized using the Adam optimizer with a weight decay of 1e-4. To learn more effective feature representations, our model is updated using the loss function $\mathcal{L}_{all}=\alpha_{c}\mathcal{L}_{con}+\alpha_{s}\mathcal{L}_{s}+\alpha_{e}\mathcal{L}_{e}$, which comprises a contractive loss $\mathcal{L}_{con}$, a scliency proposal loss $\mathcal{L}_{s}$, and a entropy loss $\mathcal{L}_{e}$.

1) The contractive loss $\mathcal{L}_{con} = -\log\left(1 - \frac{\exp(\boldsymbol{S}^{i} \odot \boldsymbol{T}^{i} / \beta)}{\sum_{\gamma \in \mathcal{B}} \exp(\boldsymbol{S}^{i} \odot T^{\gamma} / \beta)}\right)$, aims to align the proposal features with moment relationships within the $i$-th batch labels. Here, $\mathcal{B}$ denotes the index set within the batch, and $\beta$ represents the scaling factor.

2) The saliency proposal loss $\mathcal{L}_{s} = L{1}(\boldsymbol{G}^{b} - \boldsymbol{S}^{b}) + L_{1}(\boldsymbol{G}^{e} - \boldsymbol{S}^{e})$ measures the L1 distance between the predicted saliency proposal set ($\boldsymbol{S}^{b}, \boldsymbol{S}^{e}$) and the ground truth time intervals ($\boldsymbol{G}^{b}, \boldsymbol{G}^{e}$). The $L_{1}$ norm $L_{1}(\cdot)$ is defined as $0.5x^2$ for $|x| < 1$ and $|x| - 0.5$, enhancing the accuracy of saliency detection.

3) The entropy loss $ \mathcal{L}_{e}= -\sum_{m=1}^{M} \log \frac{\sum_{v \in \boldsymbol{V}_{m}^{\text {pos }}} \exp ((\boldsymbol{O}(v_{m}) / \beta_{e})}{\sum_{v \in\left(\boldsymbol{V}_{m}^{\text {pos }} \cup \boldsymbol{V}_{m}^{\text {neg }}\right)} \exp (\boldsymbol{O}(v_{m}) / \beta_{e})}$ is designed to diminish the saliency scores of non-corresponding pairs, where $\beta_{e}$ represents the scaling parameter.
\begin{table}[]
\centering
\caption{Performance comparison on QVHighlights.}
\label{your-label}
\renewcommand\arraystretch{1.1}
\begin{tabular}{lcccccccc}
\toprule[1pt]
\multicolumn{1}{c}{\multirow{3}{*}{Method}} & \multicolumn{4}{c}{test}                                    & \multicolumn{4}{c}{val}                          \\ \cline{2-9} 
\multicolumn{1}{c}{}                        & \multicolumn{2}{c}{R1} & \multicolumn{2}{c}{mAP}            & \multicolumn{2}{c}{R1} & \multicolumn{2}{c}{mAP} \\ \cline{2-9} 
\multicolumn{1}{c}{}                        & @0.5       & @0.7      & @0.75 & \multicolumn{1}{c|}{Avg.}  & @0.5       & @0.7      & @0.75      & Avg.       \\ \hline
MCN$^{\mathrm{ICCV17}}$ \cite{MCN}                                 & 11.41      & 2.72      & 8.22  & \multicolumn{1}{c|}{10.67} & -          & -         & -          & -          \\
XML$^{\mathrm{ECCV20}}$ \cite{XML}                                & 41.83      & 30.35     & 31.73 & \multicolumn{1}{c|}{32.14} & -          & -         & -          & -          \\
XML+$^{\mathrm{ECCV20}}$ \cite{XML}                               & 46.69      & 33.46     & 34.67 & \multicolumn{1}{c|}{34.90} & -          & -         & -          & -          \\
Moment-DETR$^{\mathrm{NIPS21}}$ \cite{Moment-DETR}                       & 52.89      & 33.02     & 29.40 & \multicolumn{1}{c|}{30.73} & 53.94      & 34.84     & -          & 32.20      \\
UMT$^{\mathrm{CVPR22}}$ \cite{UMT}                               & 56.23      & 41.18     & 37.01 & \multicolumn{1}{c|}{36.12} & 60.26      & 44.26     & -          & 38.59      \\
QD-DETR$^{\mathrm{CVPR23}}$ \cite{QD-DETR}                            & 62.40      & 44.98     & 39.88 & \multicolumn{1}{c|}{39.86} & 62.68      & 46.66     & 41.82      & 41.22      \\
UniVTG$^{\mathrm{ICCV23}}$ \cite{UniVTG}                             & 58.86      & 40.86     & 38.20 & \multicolumn{1}{c|}{35.47} & 60.96      & 59.74     & -          & -          \\
EaTR$^{\mathrm{ICCV23}}$ \cite{EaTR}                               & -          & -         & -     & \multicolumn{1}{c|}{-}     & 61.36      & 45.79     & 41.91      & 41.74      \\ \hline
\rowcolor{gray!10}\textbf{SpikeMba (Ours)}            & \textbf{64.13}      & \textbf{49.42}     & \textbf{43.67} & \multicolumn{1}{c|}{\textbf{43.79}} & \textbf{65.32}      & \textbf{51.33}     & \textbf{44.96}      & \textbf{44.84}      \\ 
\bottomrule[1pt]
\end{tabular}
\end{table}

\begin{table}
\centering
\caption{Performance comparison (Recall@K) on TACoS and Charades-STA.}
\label{your-label}
\renewcommand\arraystretch{1.1}
\begin{tabular}{lcccccccc}
\toprule[1pt]
\multicolumn{1}{c}{\multirow{2}{*}{Method}} & \multicolumn{4}{c}{TACoS}                          & \multicolumn{4}{c}{Charades-STA} \\ \cline{2-9} 
\multicolumn{1}{c}{}                        & @0.3 & @0.5 & @0.7 & \multicolumn{1}{c|}{mIoU}  & @0.3  & @0.5  & @0.7  & mIoU  \\ \hline
2D-TAN$^{\mathrm{AAAI20}}$ \cite{2d-TAN}                                      & 44.01 & 27.99 & 12.92 & \multicolumn{1}{c|}{27.22} & 58.76  & 46.02  & 27.5   & 41.25 \\
Moment-DETR$^{\mathrm{NIPS21}}$ \cite{Moment-DETR}                                 & 37.97 & 24.67 & 11.97 & \multicolumn{1}{c|}{25.49} & 65.83  & 52.07  & 30.59  & 45.54 \\
DRN$^{\mathrm{CVPR20}}$ \cite{drn}                                         & -     & -     & -     & \multicolumn{1}{c|}{-}     & -      & 53.09  & 31.75  & -     \\
QD-DETR$^{\mathrm{CVPR23}}$ \cite{QD-DETR}                                     & -     & -     & -     & \multicolumn{1}{c|}{-}     & -      & 57.31  & 32.55  & -     \\
LLaViLo$^{\mathrm{ICCV23}}$ \cite{LLAVILO}                                     & -     & -     & -     & \multicolumn{1}{c|}{-}     & -      & 55.72  & 33.43  & -     \\
UniVTG$^{\mathrm{ICCV23}}$ \cite{UniVTG}                                      & 51.44 & 34.97 & 17.35 & \multicolumn{1}{c|}{33.6}  & 70.81  & 58.01  & 35.65  & 50.1  \\ \hline
\rowcolor{gray!10}\textbf{SpikeMba (Ours)}           & \textbf{51.98} & \textbf{39.34} & \textbf{22.83} & \multicolumn{1}{c|}{\textbf{35.81}} & \textbf{71.24}  & \textbf{59.65}  & \textbf{36.12}  & \textbf{51.74} \\ \bottomrule[1pt]
\end{tabular}
\end{table}

\section{Experiment}
To validate our model's effectiveness, we compared it with current state-of-the-art video temporal grounding methods for moment retrieval on the QVHighlights \cite{Moment-DETR}, Charades-STA \cite{Charades-STA}, and TACoS \cite{TACoS} datasets, as well as highlight detection on the QVHighlights \cite{Moment-DETR}, TVSum \cite{tvsum}, and Youtubehl \cite{youtubehl} datasets. In this section, we delve deeper into the details, discussing the differences between various methods and our model. \textbf{We further demonstrate the dataset details and evaluation metrics in Appendix A.1 for better understanding.}

\subsection{Comparison with State-of-The-Art Methods}
In Table 1, we present our comparisons with state-of-the-art methods for joint moment retrieval and highlight detection tasks on QVHighlights. To ensure a fair comparison, we report numbers for both the test and validation splits. SpikeMba outperforms notable methods like QD-DETR, UMT, and Moment-DETR, achieving a Recall of 64.13\% at IoU 0.5 and 49.42\% at IoU 0.7 on the test dataset. Particularly, SpikeMba's average mAP score of 43.79 on the test dataset marks a significant improvement over the previous QD-DETR with an average mAP of 39.86. On the validation dataset, SpikeMba achieving a Recall of 65.32\% at IoU 0.5 and 51.33\% at IoU 0.7, with an average mAP of 44.84, highlighting its robustness and effectiveness in temporal video grounding tasks.

For moment retrieval results for the TACoS and Charades-STA datasets are shown in Table 2. On the TACoS dataset, SpikeMba shows remarkable improvement with Recalls of 51.98\% at IoU=0.3, 39.34\% at IoU=0.5, and 22.83\% at IoU=0.7, and a mean IoU (mIoU) score of 35.81. These results significantly surpass those of the previously SOTA model, UniVTG, which achieved lower scores across these metrics. This advancement highlights SpikeMba's superior ability to effectively understand and align multimodal inputs, especially in the context-rich and varied narratives of TACoS. In the Charades-STA dataset, SpikeMba continues to establish new SOTA, with Recalls of 71.24\% at IoU=0.3, 59.65\% at IoU=0.5, and 36.12\% at IoU=0.7, and an mIoU score of 51.74. These results validate the model's superiority over competitors like UniVTG and Moment-DETR and highlight its robustness and adaptability across various video content types and complexities.

We also demonstrate the highlight detection accuracy on TVsum and Youtube-hl datasets in Table 3. SpikeMba achieves an average score of 85.8 on the TVsum dataset, outperforming all compared methods. Notably, SpikeMba's scores in GA (93.0) and PR (91.5) highlight its exceptional ability to accurately identify and summarize the most relevant video content. In the Youtube-hl dataset, SpikeMba maintains a consistent lead with an overall average score of 75.5. SpikeMba's balanced performance across diverse activities, with scores ranging from 74.3 to 75.5, illustrates its adaptability and effectiveness in handling a wide range of video content.

\begin{table}
\centering
\caption{Performance comparison on TVsum and Youtube-hl.}
\label{your-label}
\renewcommand\arraystretch{1.1}
\begin{tabular}{lllllllllll}
\hline
\multicolumn{1}{c}{\multirow{2}{*}{Method}} & \multicolumn{5}{c}{TVsum}                                                                 & \multicolumn{5}{c}{Youtube-hl}                                                              \\ \cline{2-11} 
\multicolumn{1}{c}{}                        & \multicolumn{1}{c}{VT} & \multicolumn{1}{c}{GA} & PK   & PR   & \multicolumn{1}{l|}{Avg.} & \multicolumn{1}{c}{Dog} & \multicolumn{1}{c}{Gym.} & \multicolumn{1}{c}{Ska.} & Ski. & Avg. \\ \hline
SL-Module$^{\mathrm{ICCV21}}$ \cite{SL-module}                                   & 86.5                   & 74.9                   & 79.0 & 63.2 & \multicolumn{1}{l|}{73.3} & 70.8                    & 53.2                     & 72.5                     & 66.1 & 69.3 \\
QD-DETR$^{\mathrm{CVPR23}}$ \cite{QD-DETR}                                     & 88.2                   & 85.6                   & 85.8 & 86.9 & \multicolumn{1}{l|}{85.0} & 72.2                    & \textbf{77.4}                     & 72.7                     & 72.8 & 74.4 \\
UniVTG$^{\mathrm{ICCV23}}$ \cite{UniVTG}                                      & 83.9                   & 89.0                   & 84.6 & 81.4 & \multicolumn{1}{l|}{81.0} & 71.8                    & 76.5                     & 73.3                     & 73.2 & 75.2 \\
MINI-Net$^{\mathrm{ECCV20}}$ \cite{mini-net}                                    & 80.6                   & 78.2                   & 78.1 & 65.8 & \multicolumn{1}{l|}{73.2} & 58.2                    & 61.7                     & 72.2                     & 58.7 & 64.4 \\
TCG$^{\mathrm{ICCV21}}$ \cite{TCG}                                         & 85.0                   & 81.9                   & 80.2 & 75.5 & \multicolumn{1}{l|}{76.8} & 55.4                    & 62.7                     & 69.1                     & 60.1 & 63.0 \\
Joint-VA$^{\mathrm{ICCV21}}$ \cite{joint-va}                                    & 83.7                   & 78.5                   & 80.1 & 69.2 & \multicolumn{1}{l|}{76.3} & 64.5                    & 71.9                     & 62.0                     & 73.2 & 71.8 \\
UMT$^{\mathrm{CVPR22}}$ \cite{UMT}                                          & \textbf{87.5}                   & 88.2                   & 81.4 & 87.0 & \multicolumn{1}{l|}{83.1} & 65.9                    & 75.2                     & 71.8                     & 72.3 & 74.9 \\ \hline
\rowcolor{gray!10}\textbf{SpikeMba (Ours)}                                   & 85.6                   & \textbf{93.0}                   & \textbf{87.1} & \textbf{91.5} & \multicolumn{1}{l|}{\textbf{85.8}} & \textbf{74.4}                    & 75.4                     & \textbf{74.3}                     & \textbf{75.5} & \textbf{75.5} \\ \hline
\end{tabular}
\end{table}

\begin{table}
\centering
\caption{Ablation study on QVHighlights.}
\label{your-label}
\renewcommand\arraystretch{1.1}
\begin{tabular}{lcccccccc}
\hline
\multicolumn{1}{c}{\multirow{3}{*}{Method}} & \multicolumn{4}{c}{test}                                    & \multicolumn{4}{c}{val}                          \\ \cline{2-9} 
\multicolumn{1}{c}{}                        & \multicolumn{2}{c}{R1} & \multicolumn{2}{c}{mAP}            & \multicolumn{2}{c}{R1} & \multicolumn{2}{c}{mAP} \\ \cline{2-9} 
\multicolumn{1}{c}{}                        & @0.5       & @0.7      & @0.75 & \multicolumn{1}{c|}{Avg.}  & @0.5       & @0.7      & @0.75      & Avg.       \\ \hline
W/o CMR              & 53.29      & 40.32     & 36.68 & \multicolumn{1}{c|}{35.46} & 58.74      & 42.14     & 35.14      & 32.38      \\
W/o SSD               & 57.81      & 44.50     & 41.61 & \multicolumn{1}{c|}{40.51} & 60.31      & 47.33     & 39.74      & 38.79      \\
W/o contrastive loss                        & 63.18      & 45.12     & 43.34 & \multicolumn{1}{c|}{41.70} & 60.53      & 50.61     & 41.41      & 40.88      \\
W/o saliency proposal loss                  & 62.95      & 44.49     & 43.01 & \multicolumn{1}{c|}{40.51} & 61.33      & 47.93     & 41.94      & 41.62      \\ \hline
\rowcolor{gray!10}Transformer (baseline)                            & 54.23      & 38.02     & 38.84 & \multicolumn{1}{c|}{34.58} & 59.67      & 42.98     & 36.47          & 33.86      \\

\rowcolor{gray!10}\textbf{Mamba+SNN (Ours) }                            & \textbf{64.13}      & \textbf{49.42}     & \textbf{44.91} & \multicolumn{1}{c|}{\textbf{44.81}} & \textbf{65.32}      & \textbf{51.33}     & \textbf{43.93}      & \textbf{43.11}      \\ \hline
\end{tabular}
\end{table}
\subsection{Ablation Study}
\textbf{Efficacy of Model Components.} The impact of our model's various components is evaluated in Table 4. The model without the CMR and SSD are denoted as "W/o CMR" and "W/o SSD", respectively. "Transformer (baseline)" represents our baseline only with transformer. The removal of CMR resulted in a significant decrease in both Recall@0.5 and mAP scores across the test and validation sets. Specifically, Recall@0.5 and mAP (Avg.) scores dropped to 53.29\% and 35.46\% on the test set, and to 58.74\% and 32.38\% on the validation set, respectively. This emphasizes the reasoner's critical role in grasping contextual semantic associations and balancing semantic relevance exploration with contextual information preservation. The baseline only with transformer caused Recall@0.5 to decline from 64.13\% and 65.32\% to 54.23\% and 59.67\% on the test and validation sets, which demonstrate the superiorities of mamba in modeling long-term dependency. These results underline the importance of each component in boosting the model's temporal grounding abilities.

\textbf{Effectiveness of Different Loss Items.} We evaluate the impact of different loss items of our model in Table 4. Exclusion of contrastive loss $\mathcal{L}_{c}$ resulted in decreased Recall@0.5 and mAP scores, highlighting its crucial role in learning effective feature representations. The scores decreased to 63.18\% (Recall@0.5) and 41.70\% (mAP Avg.) on the test set, and slightly improved performance was observed on the validation set. Without saliency proposal loss $\mathcal{L}_{s}$ also negatively impacted the model's performance, particularly evident in the marginal decrease in mAP scores. This loss is crucial for refining the model's ability to accurately predict and prioritize salient proposals.
\begin{figure}
    \centering
        \includegraphics[width=0.32\textwidth]{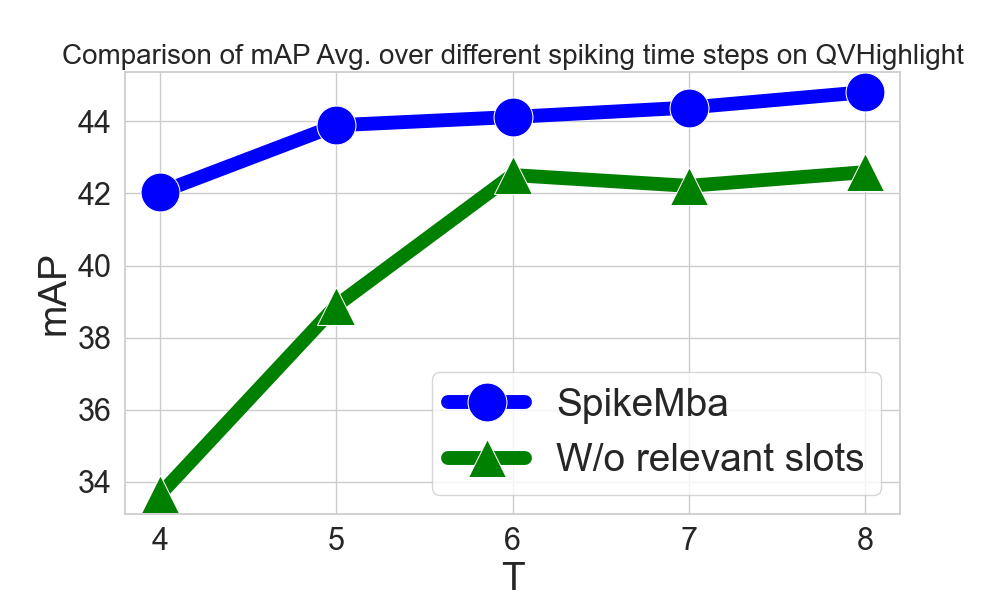}
        \includegraphics[width=0.32\textwidth]{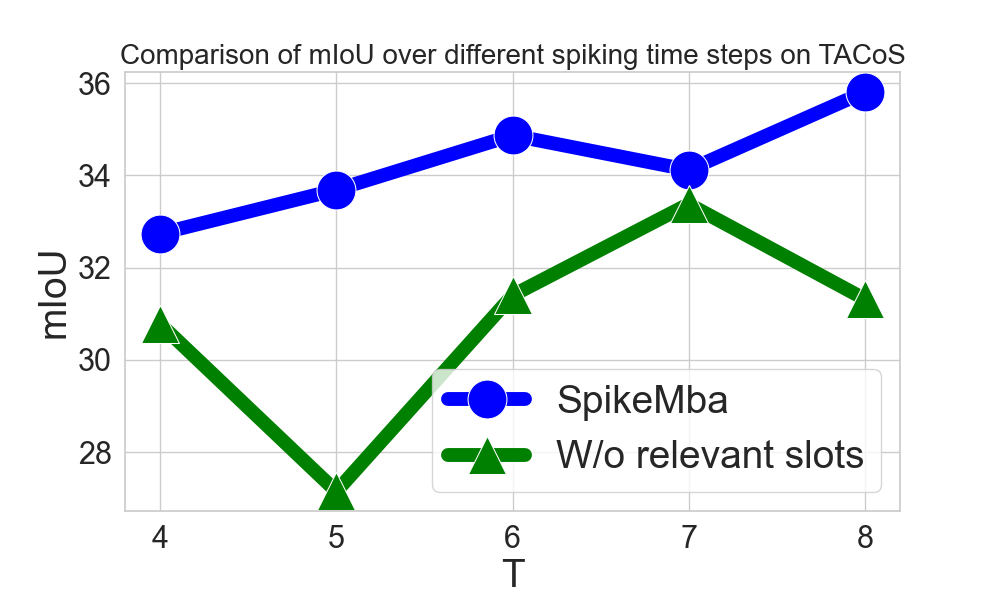}
        \includegraphics[width=0.32\textwidth]{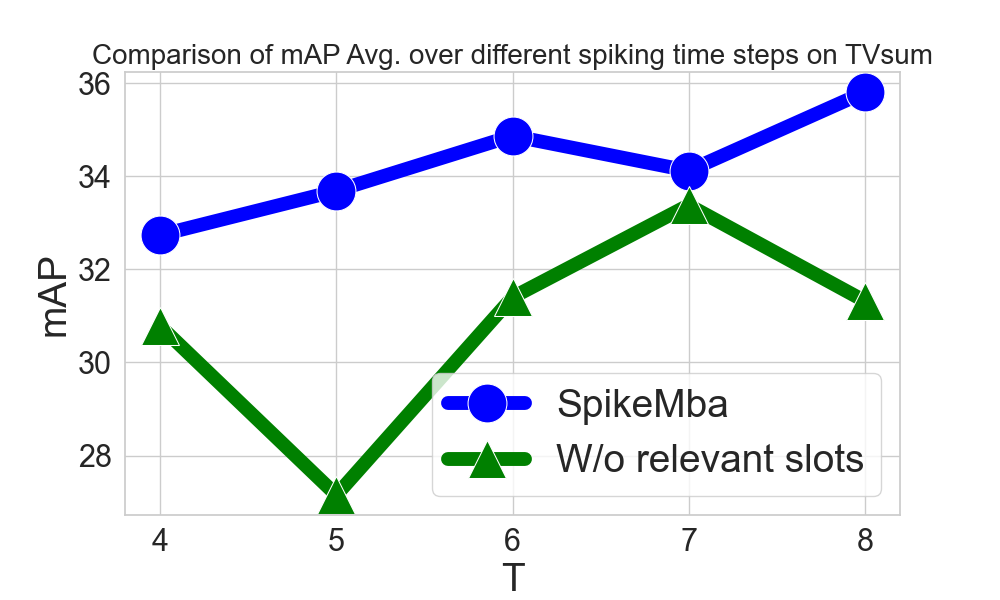}
    \caption{The ablation study of different spiking time step.}
    \label{fig:pdfs}
\end{figure}
\begin{figure}
    \centering
        \includegraphics[scale=0.5]{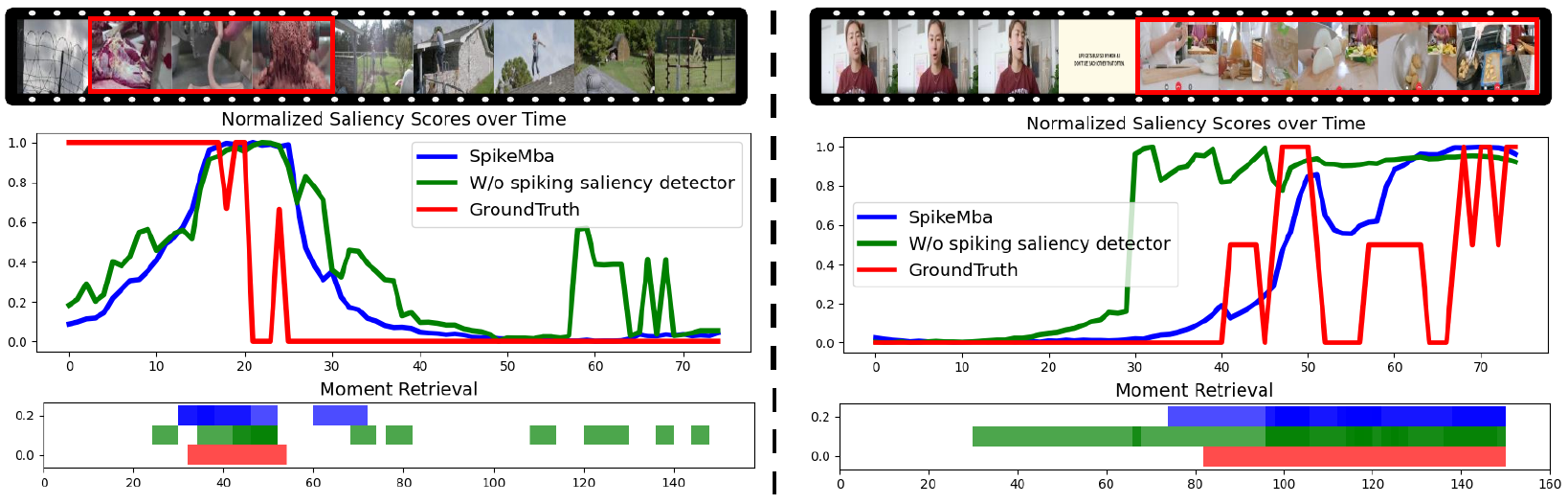}
    \caption{The qualitative results of proposed methods.}
    \label{fig:pdfs}
\end{figure}

\textbf{Effectiveness of Different Spiking Time Step and Relevant Slots.} We demonstrate the ablation study of the effectiveness of the number of spiking time steps (T) with and without the relevant slots on the model performance in Fig. 3. SpikeMba shows superior performance at higher spiking time steps, highlighting the spiking mechanism and relevant slots' importance in temporal video grounding tasks. For instance, in the QVHighlight dataset, increasing spiking time steps from T=4 to T=8 consistently improved the model's mAP (Avg.) performance. SpikeMba shows a notable improvement, from 42.03\% at T=4 to 44.81\% at T=8. In contrast, the variant without relevant slots also improves at T=8, indicating the added value of relevant slots in enhancing performance as the complexity of temporal information processing increases.

\textbf{Qualitative Results.} We illustrate some examples of qualitative results with and without the SSD in Fig. 4. The model with SSD improves moment retrieval accuracy significantly across various challenging scenarios, demonstrating accuracy in complex scenes where multiple events occur. The model demonstrates increased sensitivity to fine-grained actions and movements, accurately identifying previously overlooked moments. Furthermore, the SSD enables the model to distinguish subtle differences in similar activities, providing a clearer distinction between overlapping or temporally adjacent events, thereby enhancing the robustness and precision of video analysis.

\section{Conclusion}
This paper presented a multi-modal spiking saliency mamba for temporal video grounding, which integrates spiking neural network (SNNs) and state space models (SSMs) to address challenges like confidence bias and long-term dependency capture. The SNN-based spiking saliency detector accurately identifies salient video segments, while relevant slots and the contextual moment reasoner effectively retain and infer contextual information. SSMs enhance selective information propagation. Experiments showed that SpikeMba consistently outperforms state-of-the-art methods, demonstrating its potential to improve accuracy and efficiency in temporal video grounding.

\textbf{Limitation}: The heterogeneous nature of outputs from SNNs and the Mamba framework poses a significant challenge, requiring more effective strategies to integrate these systems and harmonize their outputs. This integration is crucial to fully leverage the complementary strengths of spatiotemporal saliency detection and long-range dependency modeling.

\small






\end{document}